\documentclass[10pt,journal,compsoc]{IEEEtran}

\ifCLASSOPTIONcompsoc
  \usepackage[nocompress]{cite}
\else
  \usepackage{cite}
\fi

\ifCLASSINFOpdf

\else

\fi

\usepackage{float}
\usepackage{amsmath}
\usepackage{graphicx}
\usepackage{caption}
\usepackage{stfloats}
\usepackage{indentfirst} 
\usepackage{multirow}
\usepackage{booktabs}
\usepackage{bigstrut}
\usepackage{hyperref}
\usepackage[table]{xcolor}
\usepackage{subfigure}
\usepackage{algorithm}
\usepackage{algorithmic}
\usepackage{amssymb}
\usepackage{longtable}
\usepackage{cite}
\usepackage{makecell}
\usepackage{soul}
\usepackage[numbers,sort,compress]{natbib}

\usepackage{amsthm,amsmath,amssymb}
\usepackage{mathrsfs}

\begin{document}

\title{IFNet: Deep Imaging and Focusing for Handheld SAR with Millimeter-wave Signals}

 \author{Yadong~Li,
         Dongheng~Zhang,
         Ruixu~Geng,
         Jincheng~Wu,
         Yang~Hu,
         Qibin~Sun,~\IEEEmembership{Fellow,~IEEE,}
         and~Yan~Chen,~\IEEEmembership{Senior Member,~IEEE}% <-this % stops a space    
\IEEEcompsocitemizethanks{\IEEEcompsocthanksitem Y. Li, D. Zhang, R. Geng, J. Wu, Q. Sun, Y. Chen are with the School
of Cyber Science and Technology, University of Science and Technology of China, Hefei 230026, China
(E-mail: yadongli@mail.ustc.edu.cn, dongheng@ustc.edu.cn, ruixu@mail.ustc.edu.cn, jinchengwu@mail.ustc.edu.cn, qibinsun@ustc.edu.cn, eecyan@ustc.edu.cn).
\IEEEcompsocthanksitem Y. Hu is with the School of Information Science and Technology, University of Science and Technology of China, Hefei 230026, China (E-mail: eeyhu@ustc.edu.cn).}% 
}

% The paper headers
%\markboth{Journal of \LaTeX\ Class Files,~Vol.~14, No.~8, August~2015}%
%{Shell \MakeLowercase{\textit{et al.}}: Bare Demo of IEEEtran.cls for Computer Society Journals}

\IEEEtitleabstractindextext{%
\begin{abstract}
Recent advancements have showcased the potential of handheld millimeter-wave (mmWave) imaging, which applies synthetic aperture radar (SAR) principles in portable settings. However, existing studies addressing handheld motion errors either rely on costly tracking devices or employ simplified imaging models, leading to impractical deployment or limited performance. In this paper, we present IFNet, a novel deep unfolding network that combines the strengths of signal processing models and deep neural networks to achieve robust imaging and focusing for handheld mmWave systems. We first formulate the handheld imaging model by integrating multiple priors about mmWave images and handheld phase errors. Furthermore, we transform the optimization processes into an iterative network structure for improved and efficient imaging performance. Extensive experiments demonstrate that IFNet effectively compensates for handheld phase errors and recovers high-fidelity images from severely distorted signals. In comparison with existing methods, IFNet can achieve at least 11.89 dB improvement in average peak signal-to-noise ratio (PSNR) and 64.91\% improvement in average structural similarity index measure (SSIM) on a real-world dataset. 
\end{abstract}

\begin{IEEEkeywords}
Millimeter-wave signal, SAR imaging, deep unfolding network.
\end{IEEEkeywords}}

% make the title area
\maketitle
\IEEEdisplaynontitleabstractindextext

\IEEEpeerreviewmaketitle

\IEEEraisesectionheading{\section{Introduction}\label{sec:introduction}}
\IEEEPARstart{M}illimeter-wave (mmWave) imaging has enormous potential in various applications, including security checks \cite{concealweapon},  human sensing \cite{mmface, rfmask, mmfer}, indoor mapping \cite{mapping1, diffradar}, and autonomous driving \cite{millipoint, auto2, dreampcd}. Its advantages of being penetrative, light-robust, and non-ionizing radiated, make it a highly promising imaging modality compared to optical cameras and X-rays \cite{digesture, millisign, pose, shuai, mobi2sense, xuan}. In addition, existing works have demonstrated the potential of mmWave imaging in detecting skin cancer \cite{srskincancer}, providing a non-intrusive approach to medical diagnosis.

Despite the attractive properties of mmWave imaging, its widespread adoption is still impeded due to the limitation of imaging resolution. With the development of large-bandwidth mmWave technology, the range resolution can be increased to about 4 cm with a 4 GHz bandwidth. However, the angle resolution, including the azimuth and elevation resolution, is inherently constrained by the aperture size, which is determined by the number of antennas and their spacing (typically half wavelength) \cite{rfimaging, wifract, ying, ying2}. Hence, existing high-resolution mmWave imaging solutions usually attempt to increase the aperture size and can be classified into two categories. 
\begin{figure}[htbp]
	\centering
	\includegraphics[scale=0.6]{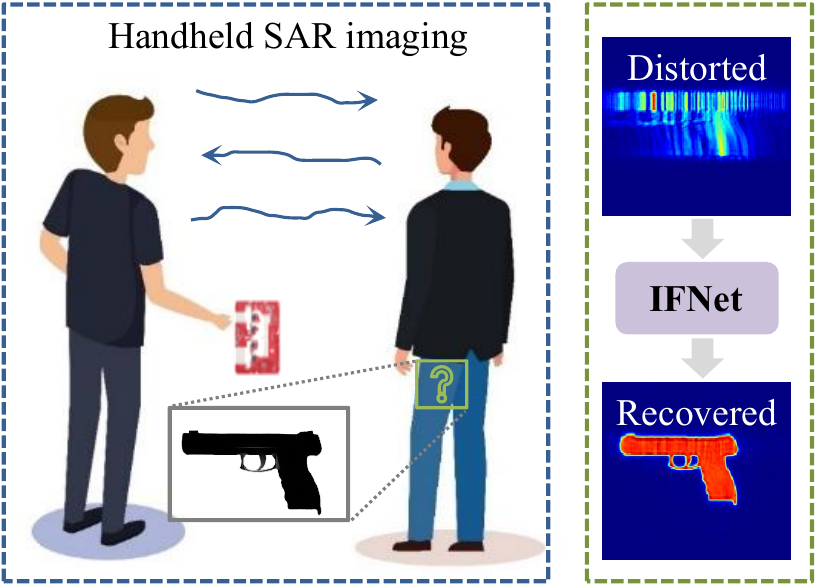}
	\caption{Handheld SAR imaging has portable applications and IFNet is designed to reconstruct images distorted by handheld motion errors.}
	\label{intro}
\end{figure} 
The first kind of technique is multi-input-multi-output (MIMO), which aims to create a larger virtual array by coordinating the locations of multiple physical antenna arrays \cite{mimo}. However, 3D high-resolution imaging usually requires hundreds of transceivers, and building such large-scale antenna arrays can be expensive and challenging. Another more cost-effective approach is synthetic aperture radar (SAR), which synthesizes large arrays by moving the physical antennas to transceive signals at different locations \cite{millipoint}. Nevertheless, it is crucial to precisely track the device's trajectory to ensure a coherent combination of received signals. Consequently, existing near-field mmWave imaging systems rely on bulky motion controllers to ensure a uniform and linear motion pattern \cite{mmface, access}, which is cumbersome and inconvenient for practical applications.

Driven by the need for compact and portable imaging systems, researchers have explored replacing mechanical scanning with handheld scanning to achieve mmWave SAR imaging \cite{ce, handheld_motioncapture, SquiggleMilli, millipcd, millicam}. One promising application of such handheld systems is security checks at train stations or airports. Currently, the inspection of suspicious threats beneath human clothes is dependent on metal detectors which can only report coarse-grained classification results, or body scanners which are built with expensive antenna arrays. In contrast, handheld SAR imaging provides a portable, cost-effective, and high-resolution solution for these applications.
However, the challenge is that manual scanning introduces significant motion errors which corrupt the phase coherency and result in severe image distortion. While pioneering work \cite{handheld_motioncapture, handheld_app1, handheld_app2} utilize motion compensation techniques to compensate for handheld motion errors, they rely on high-precision motion tracking systems because the phase of mmWave signal is sensitive to sub-millimeter-level distance variations. While other approaches employ cost-effective and coarse-grained tracking methods, they still need reference targets \cite{ce}, iterative optimization \cite{mmsight}, or time-consuming accumulation \cite{access265} to generate high-fidelity images.

Traditional model-based autofocus techniques \cite{autofocus}, aiming at estimating phase errors from received signals with signal processing, have been widely adopted in airborne SAR. For example, based on the correlation between phase errors and image quality, prior work was proposed to compensate for the phase error by optimizing certain image criteria, such as entropy \cite{entropy1, entropy2} or sharpness \cite{sharpness1, sharpness2}. While these methods have explicit theoretical foundations, they would face challenges in near-field 3D mmWave imaging because the optimization difficulty and computational cost for a large-scale array (usually tens of thousands of elements) are unacceptable. To address these issues, recent studies \cite{srgan, amgan} shift their attention to learning-based methods, demonstrating huge success in image processing, for handheld SAR imaging. However, since these methods are based on neural networks designed for processing real-valued RGB images, they only take the intensity of complex-valued SAR images as the input while discarding the phase information. In other words, the phase error estimation problem is treated as an image super-resolution problem. Such simplification fails to exploit the signal phase critical for motion compensation and does not consider the characteristics of phase error estimation. Consequently, these black-box networks encounter difficulties in achieving promising imaging performance and usually have poor generalization across targets with different sizes or shapes.

This paper introduces IFNet, which bridges the gap between signal processing models and deep neural networks to achieve robust imaging and focusing for handheld mmWave systems. 
Recognizing the complementary characteristics of model-based methods and learning-based approaches, we propose to integrate the advantages of different methodologies rather than simply relying on a single model. Specifically, we first formulate the problem of phase error compensation in handheld imaging as an optimization task. The formulated model further incorporates multiple priors considering the characteristics of mmWave images and handheld phase errors to better regularize the optimization. On this basis, to effectively obtain the optimal solution, we unfold the iterative optimization process into a deep neural network, harnessing the powerful and efficient mapping ability of learning-based methods. 
The network performs image formation
and phase error compensation with separate modules, which fully exploits the inherent characteristics of phase distribution and generates reconstructed high-fidelity images.  By mapping the signal processing model into a deep learning framework, our approach combines the advantages of both methodologies, resulting in improved interpretability and generalizability. 
Experimental results show that IFNet successfully recovers images severely distorted by handheld motion errors. Compared with baseline methods, IFNet can achieve at least 11.89 dB improvement in average peak signal-to-noise ratio (PSNR) and 64.91\% improvement in average structural similarity index measure (SSIM) on a real-world dataset.

This paper is a comprehensive extension of our previous work \cite{ifnet} by including more analysis of the challenges in handheld SAR imaging, comparing our approach with more related studies, conducting experiments with additional datasets, and exploring the impact of different design choices.   The contributions of this paper are summarized as
follows:  

(1) We propose, to our knowledge, the first imaging and focusing model that integrates signal processing approaches into the deep learning framework to perform robust phase error estimation for handheld SAR imaging.

(2) We conduct extensive experiments with various handheld scanning patterns and imaging targets, demonstrating that our approach outperforms existing methods in combating handheld motion errors and reconstructing unseen objects.

(3) We will release the implementation of IFNet and the collected dataset to facilitate the research community. 

The rest of the paper is organized as follows: Section \ref{relatedwork}
provides a comprehensive review of related works. Section \ref{preliminary} discusses the fundamental principles and specific challenges in handheld SAR imaging. Section \ref{method} introduces the detailed design of IFNet. Section \ref{experiments} demonstrates experimental validation
and comparisons with existing approaches. Section \ref{conclusion} concludes the paper.

\section{Related work}\label{relatedwork}
% In this section, we provide a comprehensive literature review including handheld mmWave imaging, autofocus, and deep unfolding network.  We focus on discussing the drawbacks of existing methods, as well as introducing the motivation and superiority of our work.

\subsection{Handheld mmWave Imaging}
Compared with traditional SAR imaging with bulky mechanical scanners, handheld SAR imaging has gained increasing attention because it is more compact and portable. Such handheld systems have the potential to enable a range of applications that require high mobility and flexibility. The key challenge of handheld SAR imaging is phase error estimation and compensation. Previous methods relied on accurate device tracking to ensure coherent signal combination \cite{handheld_motioncapture}. However, since the wavelength of the mmWave signal is extremely short, even sub-mm-level motion error can significantly disturb the phase history. Hence, to achieve reliable tracking for mmWave devices, it is necessary to employ special motion capture systems with sub-mm-level tracking accuracy, which usually costs more than \$ 40,000. To tackle this issue, prior work leveraged low-cost tracking techniques with lower accuracy. Nevertheless, they still need to handle tracking errors by time-consuming accumulation \cite{access265} or iteration \cite{mmsight}. One recent work \cite{ce} proposed to approximate the phase error of the unknown imaging target with another known reference target, and estimate the handheld phase error by optimizing the point spread function. However, this approach assumes there are suitable point scatters as the reference, which cannot be satisfied in every scenario.

In addition to model-based methods, researchers have also leveraged learning-based approaches for handheld mmWave imaging. Owing to the efficient and strong mapping capability of deep neural networks, learning-based methods can achieve better imaging performance compared with traditional signal processing techniques. However, existing approaches mainly borrow networks designed for RGB images which only have intensity. For instance, some researchers \cite{srgan} proposed to employ an efficient convolutional neural network for freehand SAR image super-resolution. Others \cite{amgan} leveraged conditional generative adversarial networks to mitigate the image artifact caused by positioning errors. While these methods reported good results on corresponding datasets, they failed to incorporate the characteristics of mmWave images (e.g. sparse and complex-valued) and the explicit model of phase error estimation.  Consequently, they face challenges in achieving robust imaging performance across different targets.

To address these issues, we design a neural network that can better model the problem of handheld mmWave imaging. Our network takes the complex-valued SAR image as the input and exploits the implicit distribution of handheld phase errors. By learning to approximate the process of signal processing-based optimization, the proposed network achieves stable and robust imaging performance.

\subsection{Autofocus}
In airborne SAR systems, the trajectory of aircraft systems might be disturbed by air currents and the positioning systems could fail to provide accurate tracking results. Consequently, autofocus techniques \cite{autofocus} have been widely studied to compensate for the phase error and recover well-focused images. Autofocus aims at directly estimating phase errors from radar echos and can be classified into three categories, namely MapDrift \cite{mapdrift1}, phase gradient autofocus \cite{pga}, and image optimization approaches \cite{entropy1, entropy2, sharpness1, sharpness2}. In the MapDrift-based approach, the synthetic aperture is divided into multiple sub-apertures, and motion parameters are estimated by correlating the formulated sub-images. This approach is not suitable for near-field handheld imaging due to the limited features in sub-images. The phase gradient autofocus-based methods rely on dominant scatters in the scene and are also not applicable because of the far-filed assumption. Finally, the image optimization autofocus approach aims to improve image quality by searching for the optimal compensated phase based on specific metrics. However, high-resolution 3D mmWave imaging requires a larger number of variables that need to be optimized, leading to increased difficulty in optimization and heavy computational cost.

In this paper, we leverage deep neural networks to address these problems and perform handheld phase error compensation. Deep neural networks have demonstrated superior performance compared with traditional signal processing or optimization methods in a wide range of tasks. By exploiting the distribution of handheld phase errors from training data, efficient and improved performance can be achieved without relying on specific assumptions. Moreover, as described earlier, our network design differs from existing networks by integrating the characteristics of mmWave images and handheld phase errors, resulting in stronger generalization across different targets.

\subsection{Deep Unfolding Network}
Deep unfolding networks have been widely applied to solve various inverse problems, such as image denoising \cite{dun_denoise}, image super-resolution \cite{dun_sr}, and compressive sensing \cite{dun_cs}. The basic design principle is to convert an iterative optimization process into a deep neural network. Each layer of the network corresponds to one iteration of the optimization. One advantage of deep unfolding networks is that they can automatically learn various parameters through training, rather than manually adjusting in traditional optimization.
Based on this principle, researchers have made significant advancements in transforming different optimization algorithms into deep neural networks, including iterative shrinkage-thresholding algorithm (ISTA) \cite{ISTAnet} and alternating direction method of multipliers (ADMM) \cite{ admmnet2}.

Similarly, our network is based on the formulated optimization model for handheld imaging and focusing. By designing the network according to the optimization steps of mmWave imaging, our model is more interpretable compared with existing handheld imaging networks that directly borrow the structures for RGB images.

\section{Preliminary and challenges}\label{preliminary}
In this section, we describe the fundamental principles of MIMO SAR imaging to provide basic knowledge of this paper. Additionally, we illustrate the challenges in handheld SAR imaging through both theoretical and empirical analysis.

\begin{figure}[htbp]
	\centering
	\includegraphics[scale=0.8]{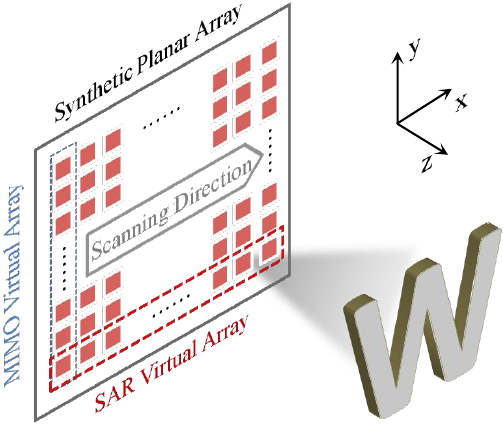}
	\caption{Imaging model of MIMO SAR.}
	\label{mimosar}
\end{figure} 

\subsection{MIMO SAR Imaging Fundamentals}\label{sarfundamental}
A 2D planar aperture is required to achieve 3D mmWave imaging. With the combination of Multi-Input Multi-Output (MIMO) technology and the SAR principle, one can synthesize such a virtual planar array with much more efficient data acquisition. As illustrated in Fig. \ref{mimosar}, an ideal planar aperture can be synthesized by horizontally moving a linear MIMO array with a mechanical scanner. Let $x$, $y$, and $z$ axis be the scanning direction, MIMO array direction, and depth direction, respectively. Suppose the transmitting antennas and receiving antennas are located at $(x_T,y_T,0)$ and $(x_R,y_R,0)$, respectively. The imaging target is represented by a set of point scatters located at $(x,y,z)$ with a reflectivity function $\sigma(x,y,z)$. 
Under the Born approximation for the scattering
field and an isotropic antenna assumption, the received signal of the imaging space $D$ can be expressed as
\begin{equation}\label{receivesignal}
	s(x_T,y_T,x_R,y_R,k) = \iiint\limits \sigma(x,y,z) e^{-jk(R_{Tx}+R_{Rx})}dxdydz 
\end{equation}
where $k = \frac{2\pi f}{c}$ and $f,c,k$ represent the temporal frequency, speed of propagation, and wavenumber, respectively. $R_{Tx}$ and $R_{Rx}$ are the distances from the transmitter and receiver to the imaging points, which can be denoted as
\begin{gather} 
	R_{Tx} = \sqrt{(x-x_T)^2+(y-y_T)^2+z^2} \\
	R_{Rx} = \sqrt{(x-x_R)^2+(y-y_R)^2+z^2}
\end{gather}

\textbf{Time-domain imaging:} The well-known time-domain back-projection algorithm (BPA) can be directly derived from Eq. \ref{receivesignal}.  Specifically, the target’s reflectivity function can be recovered from the receiving signal as follows:

\begin{multline}\label{bpa}
    \begin{aligned}
            	\sigma(x,y,z)  &= \iiint\limits s(x_T,y_T,x_R,y_R,k) \\ &\times e^{jk(R_{Tx}+R_{Rx})}dx_Tdy_Tdx_Rdy_Rdk 
    \end{aligned}
\end{multline}

\textbf{Frequency-domain imaging:}
Range migration algorithm (RMA) \cite{rma}, the most popular frequency-domain imaging method, is based on monostatic sampling schemes. Hence, a multistatic-to-monostatic conversion \cite{access} is required for near-field MIMO SAR imaging with a large aperture, as shown in the following:
\begin{equation}
	\begin{aligned}
		\hat{s}(x',y',k) &= s(x_T,y_T,x_R,y_R,k)e^{-jk(\frac{{d_x}^2+{d_y}^2}{4z_0})} \\
		&=  \iiint\limits \sigma(x,y,z) e^{-j2kR}dxdydz 
	\end{aligned}   %%%%  target (x,y,z) or (x,y,Z_1)
\end{equation}
where $(x',y',0)$ is the phase center of the transmitter at $(x_T , y_T , 0)$ and the receiver at $(x_R, y_R, 0)$.  $d_x$ and $d_y$ are the distances between the transmitter and
receiver along $x$ and $y$ axes, respectively. $z_0$ is the depth of a reference point located at $(x_0,y_0,z_0)$. $R$ is the distance from the phase center to the target points.

We consider the reconstruction of a 2D imaging plane located at $z=r$ and $k=k_r$, then the 2-D receiving signal can be rewritten as 
\begin{equation}\label{sphericalwave}
	\hat{s}(x',y',k_r) = \iint\limits \sigma(x,y,r) e^{-j2k_rR}dxdy
\end{equation}
The exponential term in the above equation is the representation of a spherical wave, which can be decomposed into 
a superposition of plane waves,
\begin{equation}\label{planewaves}
	e^{-j2k_rR} = \iint e^{-jk_x(x-x')-jk_y(y-y')-jk_zr }dk_xdk_y
\end{equation}
where $k_z=\sqrt{4k_r^2-{k_x}^2-{k_y}^2}$. $k_x$, $k_y$ and $k_z$ are the Fourier transformation variables corresponding to $x$,$y$ and $z$, respectively. 
Substituting Eq. \ref{planewaves} into Eq. \ref{sphericalwave} yields 
%\begin{equation}
%	\begin{aligned}
%		\hat{s}(x',y',r) &= \iint\limits \sigma(x,y,r) \times 
%		\{\iint e^{-jk_x(x-x')-jk_y(y-y')-jk_zr }dk_xdk_y\} dxdy \\
%		&= \iint \{\iint\limits \sigma(x,y,r)e^{-j(k_xx+k_yy)}dxdy\} \times 
%		e^{-j(k_xx'+k_yy'+k_zr)}dk_xdk_y \\
%		&= {FT_{2D}}^{-1}[FT_{2D}[\sigma(x,y,r)]e^{jk_zr}]
%	\end{aligned}
%\end{equation}

\begin{multline}
	\begin{aligned}
		\hat{s}(x',y',k_r) &= \iint\limits \sigma(x,y,r) \\
		&\times 
		\{\iint e^{-jk_x(x-x')-jk_y(y-y')-jk_zr }dk_xdk_y\} dxdy \\
		&= \iint \{\iint\limits \sigma(x,y,r)e^{-j(k_xx+k_yy)}dxdy\} \\
		&\times 
		e^{-j(k_xx'+k_yy'+k_zr)}dk_xdk_y \\
		&= {FT_{2D}}^{-1}[FT_{2D}[\sigma(x,y,r)]e^{jk_zr}]
	\end{aligned}
\end{multline}
where $FT_{2D}$ and ${FT_{2D}}^{-1}$ represents the 2-D Fourier transformation
and the 2-D inverse Fourier transformation, respectively. 
Finally, the 2-D SAR image reconstruction using RMA can be formulated as
% \begin{equation}\label{rma}
% 	\sigma(x,y,k_r)  = {FT_{2D}}^{-1}[FT_{2D}[ \hat{s}(x',y',r)]e^{jk_zr}]
% \end{equation}
\begin{equation}\label{rma}
	\sigma(x,y,r)  = {FT_{2D}}^{-1}[FT_{2D}[ \hat{s}(x',y',k_r)]e^{jk_zr}]
\end{equation}

\textbf{Time-domain imaging v.s. frequency-domain imaging:} From Eq. \ref{bpa} and Eq. \ref{rma}, we can find that BPA is computationally intensive while RMA, which is based on FFT, is much more efficient. On the other hand, BPA is suitable for arbitrary paths but RMA requires a linear and uniform motion. In addition, both of these two algorithm need to know the accurate positions of devices. Since a practical handheld imaging system requires a high-efficiency imaging algorithm, in this paper, we will focus on RMA imaging under non-uniform and non-linear motion. However, we will also use BPA to illustrate the impact of localization error in the following sections.

\begin{figure}[H]
	\centering 
	\subfigbottomskip=2pt 
	\subfigcapskip=-2pt 
	\subfigure[Ground truth without motion errors.]{
		\includegraphics[width=0.35\linewidth]{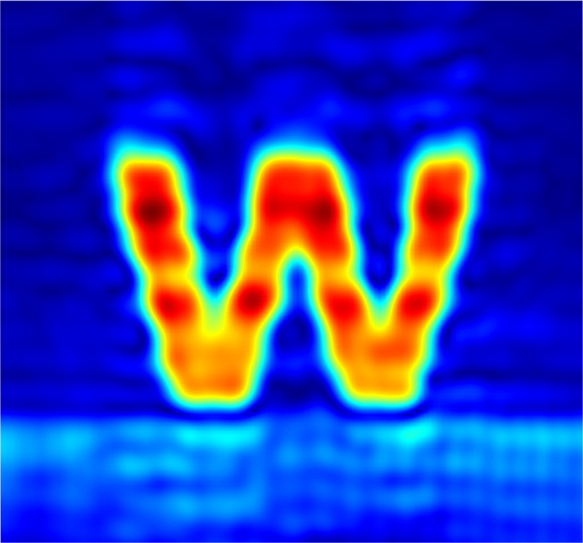}}
        \quad
	\subfigure[Adding simulated errors with std=0.3 mm.]{
		\includegraphics[width=0.35\linewidth]{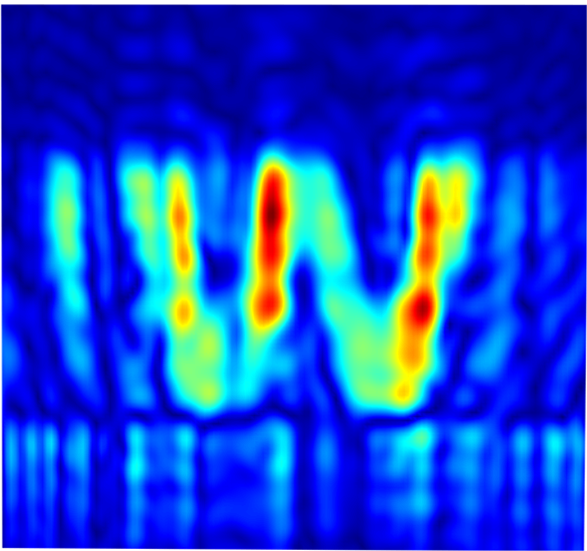}}
	  \\
	\subfigure[Adding simulated errors with std=0.5 mm.]{
		\includegraphics[width=0.35\linewidth]{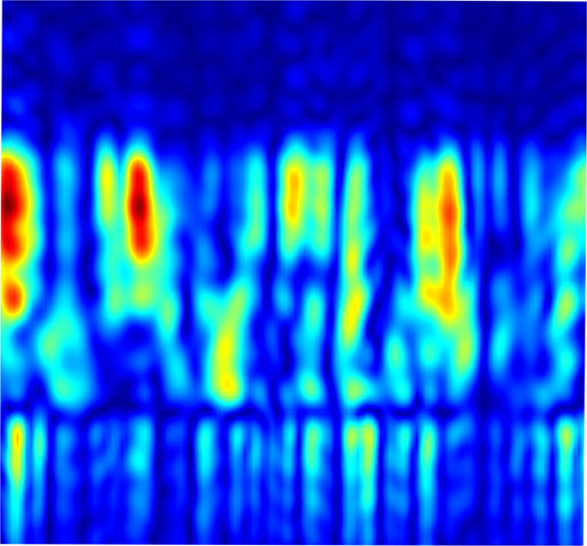}}
	\quad
	\subfigure[Adding measured errors of a tracking camera.  ]{
		\includegraphics[width=0.35\linewidth]{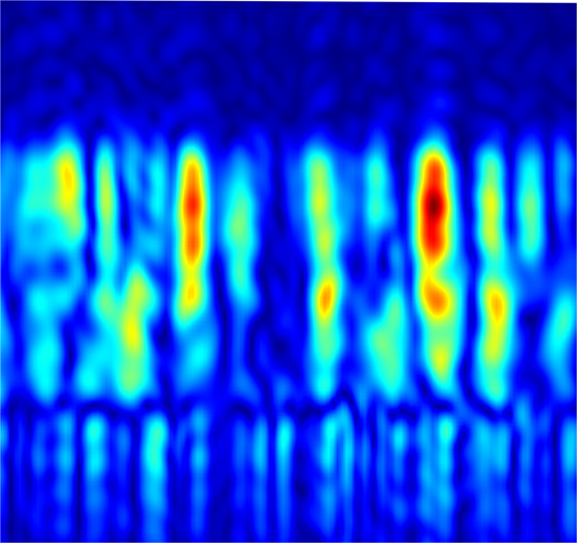}}
	\caption{Impact of handheld motion errors. Even sub-mm scale motion errors can lead to significant image distortion.}
\end{figure}\label{motion_error_analysis}

\subsection{Analysis of Handheld Motion Errors}
Conventional mmWave SAR imaging requires motion controllers to move the device along a predefined path, which is usually linear and uniform motion. The reason for this requirement is to make sure the received signals at different locations are coherent and can be combined to achieve higher angle resolution. However, in handheld imaging settings, it is impractical to expect users to complete such an ideal motion with their arms and hands. Consequently, the phase coherency of received signals will be corrupted by motion errors and the formulated images will be severely distorted.

To illustrate the impact of motion errors on the image quality, we simulate different levels of motion errors for mmWave SAR imaging. As demonstrated in our previous work \cite{ce}, the motion error along the range axis has the most prominent influence. Therefore, we first collect ideal SAR data with motion controllers. Then, we add different statistical motion errors, with a standard deviation of 0.3 mm and 0.5 mm to the range axis. From the imaging results shown in Fig. \ref{motion_error_analysis}, we can observe that motion errors with a standard deviation larger than 0.5 mm will cause severe image distortion. This is because the phase of mmWave signals is sensitive to sub-millimeter scale motion, and even small motion deviations can lead to significant phase errors in the received signals. 

To address the handheld motion errors, existing work utilized motion capture systems with a sub-millimeter tracking accuracy, which is however too expensive for consumer applications. Recent studies also leverage mobile localization methods such as the Intel RealSense tracking camera. Despite being more affordable, its tracking accuracy, especially along the depth axis, fails to meet the stringent requirement of handheld mmWave imaging. Fig. \ref{motion_error_analysis}(d) illustrates the defocused imaging result when adding motion errors measured with the Intel RealSense tracking camera.
Consequently, prior works rely on reference targets, time-consuming accumulation, or iterative registration to obtain reasonable imaging results. 

Based on the above discussion, the research focus of our work is to efficiently estimate and compensate for the handheld phase errors. In what follows, we will elaborate on the formulated imaging model and network design, which can restore high-fidelity images from severely distorted input.

\section{Method}\label{method}
This section provides a detailed description of our methodology. Firstly, we introduce our model for handheld imaging and focusing with multiple priors. On this basis, we demonstrate how this optimization model is unfolded into a deep neural network. 
\vspace{-0.4cm}
\subsection{Imaging and Focusing with Sparse and Deep Prior} \label{signalmodel}
The range migration algorithm (RMA) \cite{rma, access}, a widely used traditional SAR imaging method, can be defined as a pair of operators including the imaging 
process $\mathcal{I}{(S,M)}$ and signal generating process $ \mathcal{G}{(\Sigma,\bar{M})}$ (i.e., inverse RMA), respectively \cite{mou}.
\begin{small}
	\begin{gather}
		\Sigma = {IFFT_{2D}}[FFT_{2D}[S]\odot M] =  \mathcal{I}{(S,M)}, \\
		S = {IFFT_{2D}}[FFT_{2D}[\Sigma]\odot \bar{M}] =  \mathcal{G}{(\Sigma,\bar{M})},
	\end{gather}
\end{small}where $S$ is the received signal of a planar array, $\Sigma$ is the target 2D imaging area.  $FFT_{2D}$ and $IFFT_{2D}$ denote the 2D Fast Fourier Transform (FFT) and 2D inverse FFT, respectively. $M$ represents a specific phase term and $\bar{M}$ is the conjugate version of $M$. $\odot$ denotes element-wise product.
In the presence of phase errors caused by handheld scanning, the signal model can be denoted as:
\begin{small}
	\begin{equation}\label{imagingmodel}
		\Phi \odot S_{\epsilon} = \mathcal{G}{(\Sigma,\bar{M})} + N,
	\end{equation}
\end{small}where $S_\epsilon$ is the phase-corrupted received signal, $\Phi$ stands for the corresponding phase compensation factor, and $N$ refers to additive white Gaussian noise.
Since a majority of objects in mmWave spatial frequency domain have a high degree of sparsity, Eq. \ref{imagingmodel} can be modeled as a sparsity-driven optimization problem by adding a regularization term $\Vert vec(\Sigma) \Vert_1$.
However, when the phase errors are significant, which is the case of handheld scanning, the sparsity constraint cannot guarantee satisfactory images. Since there is no prior information about the phase compensation matrix $\Phi$, it usually results in unreliable handheld phase error estimation. 

To reduce the manual scanning time and improve system efficiency, we propose to synthesize a planar array by manually moving a linear array. Hence, the handheld scanning has a certain motion pattern (a near-straight line), which means that the phase error is also subject to a specific distribution rather than completely random. Hence, prior information about $\Phi$ can be utilized to acquire stable and robust focusing results. However, handcrafted regularization is not applicable because the distribution characteristics of handheld phase error are unknown. Inspired by the success of deep neural networks, we formulate the regularization term of $\Phi$ as a deep prior, which can be learned from data both effectively and automatically.
Therefore, the problem of joint handheld SAR imaging and focusing can be formulated as the problem of minimizing the following objective function:
\begin{small}
	\begin{gather} \label{costfunction}
		\hat{\Phi}, \hat{\Sigma} = \mathop{\arg\min}_{\Phi,\Sigma} \frac{1}{2} {\Vert \Phi \odot S_{\epsilon} -\mathcal{G}{(\Sigma,\bar{M})} \Vert_2}^2 + \lambda \Vert vec(\Sigma) \Vert_1 + \gamma J(\Phi), 
	\end{gather}
\end{small}where $\lambda$ is the weighting parameter indicating the
strength of the regularization term $\Vert vec(\Sigma) \Vert_1$.  $J(\Phi) $ is the regularization term of $\Phi$ and $\gamma$ is the corresponding weighting factor. Eq. \ref{costfunction} can be solved by iteratively
alternating between image formation (i.e., imaging part) and phase error compensation (i.e., focusing part) using a coordinate descent technique. Specifically, in the imaging part of each iteration, the cost function is minimized by updating the image $\sigma$ with iterative shrinkage thresholding algorithm (ISTA):
\begin{small}
	\begin{gather}
		R^{(k)} = \Sigma^{(k-1)} - \mu \mathcal{I}\left(\mathcal{G}{(\Sigma^{(k-1)},\bar{M})}-\Phi^{(k-1)} \odot S_{\epsilon}^{(k-1)} \right), \label{gdi}	\\
		\Sigma^{(k)} = {prox}_{\lambda,\Vert  \Vert_1}(R^{(k)}) = soft(R^{(k)},\lambda), \label{pmi}
	\end{gather}
\end{small}where $\mu$ denotes the updating step size, $prox$ represents the proximal mapping operator, and $soft$ is the soft thresholding function.
In the focusing part, the phase compensation factor $\Phi$ is estimated given the updated image as follows:
\begin{small}
	\begin{gather}
			V^{(k)} = \Phi^{(k-1)} - \rho {S_{\epsilon}^{(k-1)}}^{\mathsf{H}} \odot \left(\Phi^{(k-1)} \odot S_{\epsilon}^{(k-1)} -\mathcal{G}{(\Sigma^{(k)},\bar{M})}\right), \label{gdf}\\
			\Phi^{(k)} = prox_{J}(V^{(k)}), \label{pmf}
	\end{gather}
\end{small}where $\rho$ denotes the updating step size. Instead of deriving the proximal mapping $prox_{J}(V^{(k)})$ as a soft thresholding function, we formulate it as a neural network structure with improved and robust focusing ability.

\begin{figure*}[htbp]
	\centering
	\includegraphics[scale=0.6]{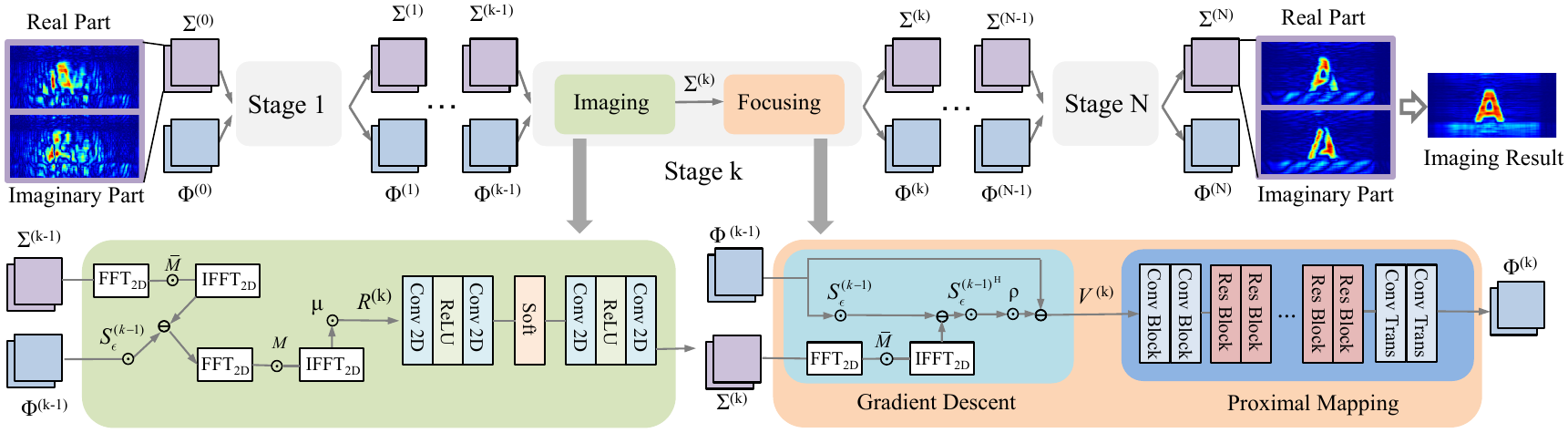}
	\caption{The architecture of IFNet. IFNet comprises multiple stages, with each stage consisting of an imaging module and a focusing module. The design of the imaging and focusing modules are based on the optimization-based signal model.}
	\label{ifnet}
\end{figure*}

\subsection{Imaging and Focusing with Deep Unfolding Network}
Tradition optimization method to solve Eq. \ref{costfunction} not only introduces a heavy computational burden due to the iterative updating but also generates low-quality images because of the limited representation ability of the handcrafted regularization. Inspired by recent advances in the interpretable deep unfolding network \cite{ISTAnet, dpn1, dpn2}, which aims to replace the handcrafted regularization with a neural network module that has powerful learning ability, we propose IFNet by mapping the update steps in Sec. \ref{signalmodel} to an end-to-end deep network structure for mmWave imaging and focusing.

\textbf{Architecture.} The architecture of IFNet is illustrated in Fig. \ref{ifnet}. Specifically, the network is composed of $K$ stages, representing a total of $K$ updating steps optimizing the objective function in Eq. \ref{costfunction}. Each stage consists of an imaging part and a focusing part that corresponds to image formation and phase error compensation. For instance, in the imaging part of $k^{th}$ stage, IFNet leverages the estimated phase compensator $\Phi^{k-1}$ to update and obtain the image $\Sigma^{k}$. If the previous estimated $\Phi^{k-1}$ is optimized towards the right direction, this step will result in better image quality because parts of the handheld phase errors have been compensated. After that, the focusing part exploits the residual distortion in image $\Sigma^{k}$ and updates $\Phi^{k-1}$ to obtain a more accurate phase compensator $\Phi^{k}$.

More specifically, since the solution to the imaging part in Sec. \ref{signalmodel} is essentially an ISTA process, we adopt the design of ISTA-Net \cite{ISTAnet} and modify the specific modules according to our task. The imaging part contains the gradient descent corresponding to Eq. \ref{gdi} and two convolutional blocks with a soft-thresholding function corresponding to Eq. \ref{pmi}. Each convolutional block consists of a convolutional layer of 8 kernels with a size of 3, a ReLU activation function, and a convolutional layer of 2 kernels with a size of 3.  
The focusing part is composed of a gradient descent step corresponding to Eq. \ref{gdf} and an encoder-decoder module corresponding to Eq. \ref{pmf}. 
The encoder-decoder module consists of two convolutional blocks with the first one having 32 convolution kernels with a size of 7 and the second one having 64 convolution kernels with a size of 3. Each convolution layer is followed by an instance normalization layer and a ReLU layer. Then, we employ multiple ResBlocks \cite{resnet} to exploit more inherent features about handheld phase error distribution, and 2 transpose convolution blocks to finally obtain the updated phase compensator. 

The main advantage of IFNet against existing networks is that it integrates the signal processing approach into the deep learning framework. As a result, the imaging and focusing process of IFNet has better interpretability and stronger generalization.

\textbf{Initialization.} The input of IFNet is the SAR image distorted by the manual scanning phase errors. Different from prior work that only utilized the amplitude of the complex images, we preserve the phase information by separating the real part and imaginary part and concatenating them as a 2-channel image. This helps the network to better capture the correlation between the images and phase errors.
To facilitate the training process, we perform min-max normalization on the real and imaginary parts, respectively. 
Additionally, the derived solution in Section 4.1 includes several hyperparameters, i.e., $\lambda,\mu,\rho$, which need to be adjusted manually in the traditional optimization process. Similar to the existing work \cite{ISTAnet}, we set all of them to be learnable parameters that can be automatically optimized by the network. 

\textbf{Loss function.}  The main optimized values of IFNet are image $\Sigma$ and phase compensator $\Phi$. However, obtaining the ground-truth handheld phase compensator $\hat{\Phi}$ is challenging as it requires sub-mm scale tracking. Therefore, we choose to optimize the network by measuring the difference between the distorted image $\Sigma$ and groud-truth image $\hat{\Sigma}$. This is because if the phase compensator is optimized towards the right direction, the distorted image will be closer to the well-focused image. Additionally, although the output of IFNet is a two-channel complex image, we transform it into an amplitude image when measuring the network loss to make the training process more robust to unstable phase variations. Consequently, the loss function of mean square error can be represented as:
\begin{gather}
    \text{MSE}(\Sigma, \hat{\Sigma}) = \frac{1}{n} \sum_{i=1}^{n}(\text{abs}(\Sigma) - \text{abs}(\hat{\Sigma})^2), \\
    \text{abs}(\Sigma) = \sqrt{(\Sigma[:,:,0])^2 + (\Sigma[:,:,1])^2} 
\end{gather}
where $n$ represents the number of training samples, $\Sigma[:,:,0]$ and $\Sigma[:,:,1]$ denote the real part and imaginary part of the SAR image, respectively.

\section{Experiments}\label{experiments}
\subsection{Implementation Details}
\textbf{Device Configuration.} To collect SAR data, we employ the TI MMWCAS-RF-EVM radar consisting of 12 transmitting antennas and 16 receiving antennas. The radar operates at 76-81 GHz and can formulate an 86-element linear array with MIMO technology. We set the radar parameters as start frequency, 77 GHz;  chirp slope, 38.5 MHz/$\mu$s; chirp duration, 40 $\mu$s; the number of ADC samples, 256; ADC sampling rate, 8 Msps. 

\textbf{Network Training.} The dimension of the input and output SAR image of the network is $256 \times 128 \times 2 $. The initial values of hyperparameters are set as $\lambda = 0.01,\mu = \rho = 0.5$. We implement the network with PyTorch and train it for 80 epochs with a learning rate of 0.0001, which starts to decay after 50 epochs until becomes 0 at the 80th epoch. 

\textbf{Baseline Methods.} To demonstrate the superiority of IFNet, we have implemented existing handheld SAR imaging networks as baselines and made a comparison. Specifically, we term these cGAN-based architectures SRGAN \cite{srgan} and AMGAN \cite{amgan}, as they have been proposed for super-resolution and artifact mitigation for handheld SAR images, respectively. 

\textbf{Evaluation Metrics.} We utilize two commonly employed metrics to assess the quality of SAR images: peak-to-noise ratio (PSNR) and structural similarity index measure (SSIM). PSNR measures the amount of noise or distortion in an image and a higher PSNR value denotes better image quality. SSIM provides a more perceptually meaningful assessment of image quality. It ranges between 0 and 1, where a value of 1 indicates perfect similarity between the two images.

\begin{figure}[htbp]
	\centering
	\includegraphics[scale=0.5]{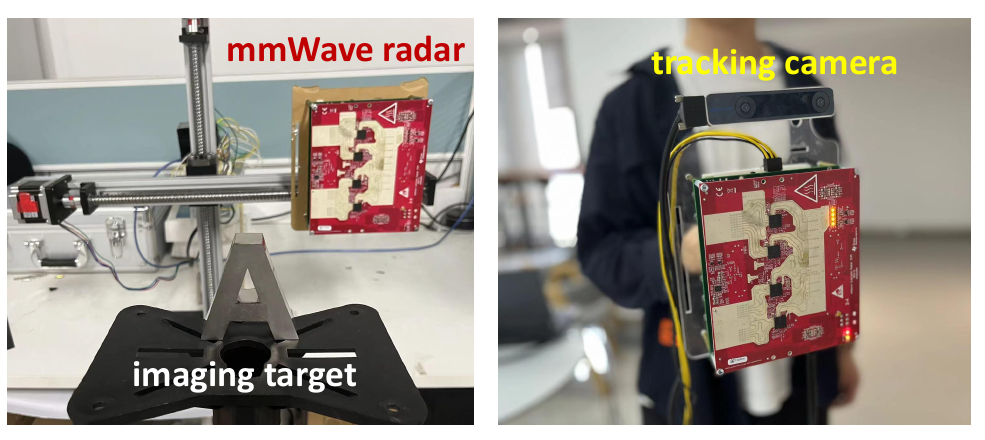}
	\caption{Collecting ground truths with mechanical scanning and handheld scanning patterns with tracking camera. }
	\label{setup}
\end{figure} 
\vspace{-0.5cm}
\subsection{Evaluation Datasets}
Large-scale and high-quality training datasets are crucial for ensuring the robustness of learning-based methods. However, there is currently no publicly available handheld SAR imaging dataset, and creating such a dataset can be highly time-consuming as it requires various imaging objects and handheld scanning patterns. Therefore, similar to the baselines \cite{srgan, amgan}, we take a hybrid approach to generate sufficient training data. Specifically, 
we first collect real measurements with mechanical scanning as ground truths, then add diverse phase errors to the measurements to simulate different handheld scanning. To evaluate the performance when testing on a large number of different objects, we also simulate SAR signals using the ShapeNet \cite{shapenet} dataset, which covers 55 common object categories with about 51,300 unique 3D models. 

\begin{figure}[htbp]
	\centering
	\includegraphics[scale=0.5]{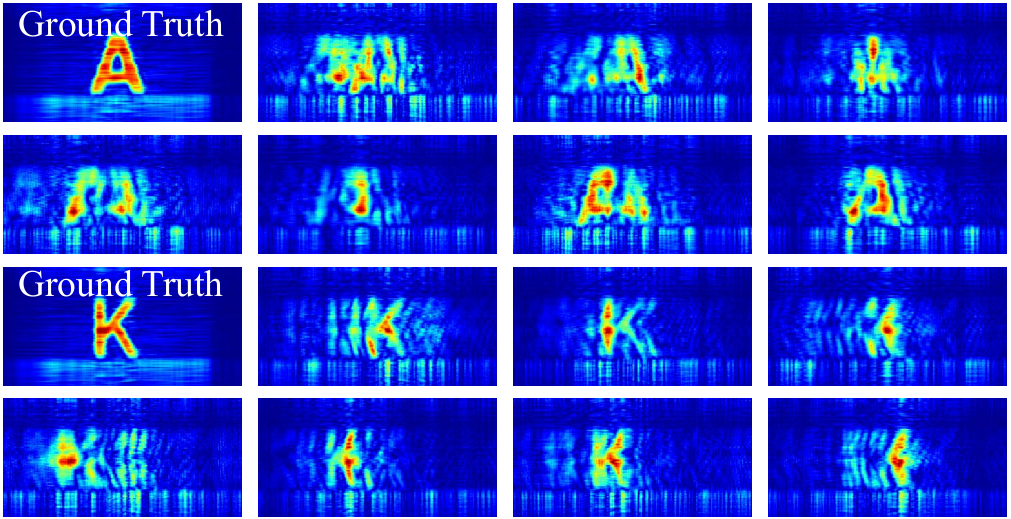}
	\caption{The ground truths of different letters and the results distorted by various handheld scanning patterns. }
	\label{dataset_letter}
\end{figure}

\subsubsection{Real measurements with 26 letters}
To generate datasets with various handheld scanning patterns, we utilize a motion controller to collect SAR signals without phase errors, then record 200 scanning trajectories performed by four volunteers using the Intel RealSense T265 tracking camera, as shown in Fig. \ref{setup}. To further diversify the dataset, we randomly select five trajectories and calculate their mean trajectory. This allows us to synthesize a large number of different handheld trajectories that can produce different phase errors. Finally, we compute the phase errors by measuring the differences between the handheld trajectories and the motion controller trajectories. These phase errors are then applied to the ground-truth images, generating distorted images that reflect handheld motion errors as shown in Fig. \ref{dataset_letter}.

The dimension of the synthetic planar array is $200 \times 86$. The distance between adjacent virtual elements along the x-axis and y-axis is 1 mm and 0.97 mm, respectively. The distance between the target and the array plane is 0.3 m, and the size of the target is 6 cm$ \times$ 6 cm.  In total, our dataset comprises 10,400 images from 26 different letters, with each letter distorted by 400 different handheld trajectories. To evaluate the performance of our model on new objects, we use 8,000 images from 20 letters for training and reserve 2,400 images from the remaining 6 letters for testing.

\begin{figure}[htbp]
	\centering
	\includegraphics[scale=0.5]{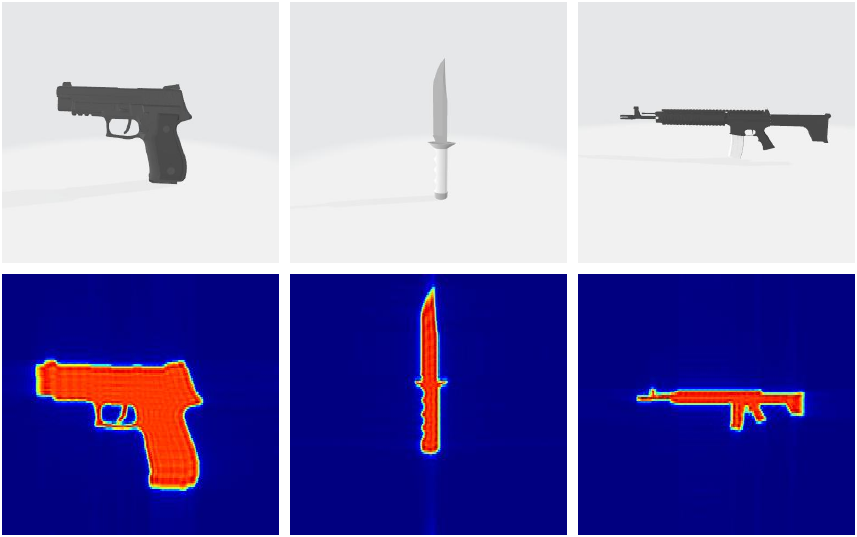}
	\caption{The 3D models in the ShapeNet dataset and their corresponding simulated SAR images. }
	\label{shapenet_dataset}
\end{figure} 

\begin{figure*}[htbp]
	\centering
	\includegraphics[scale=0.6]{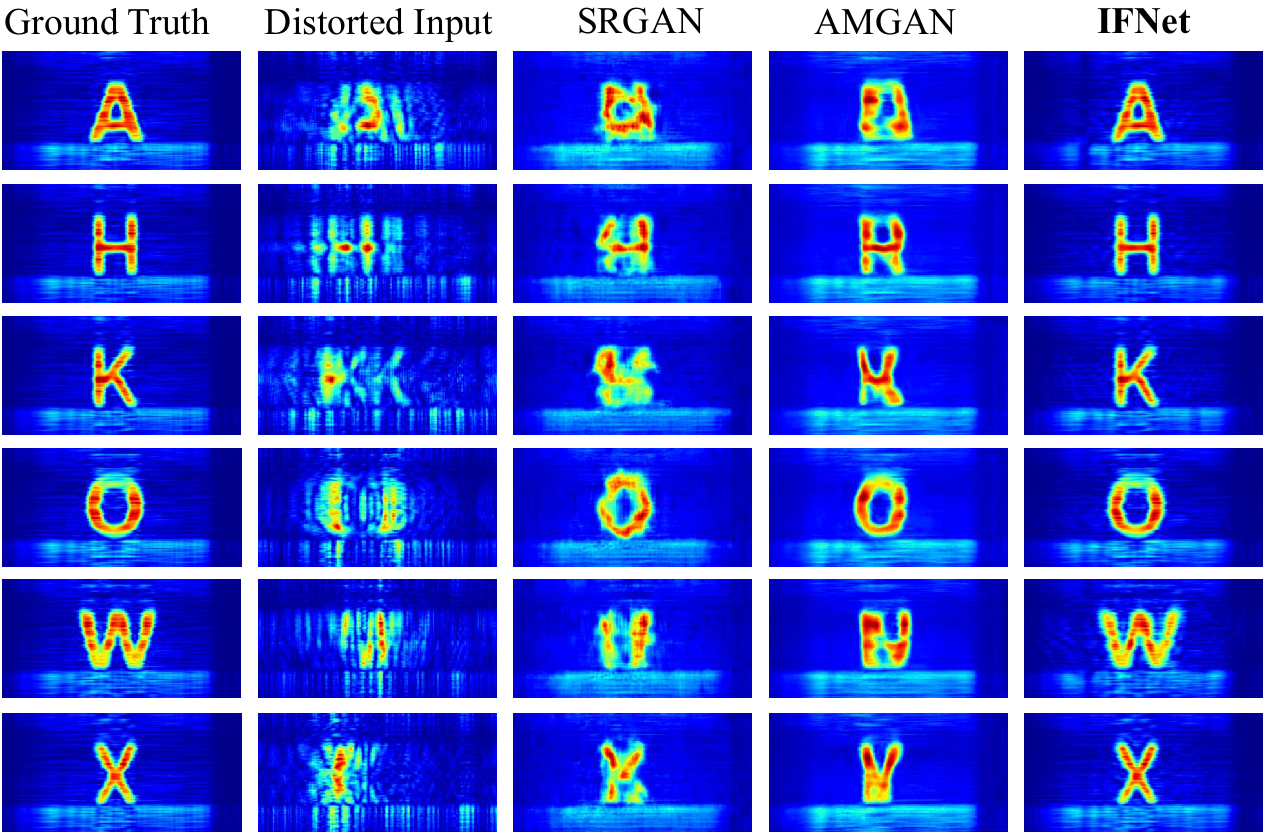}
	\caption{Qualitative comparison with 26 letters from real measurements. The testing letters are not included in the training set. Even for severely distorted images,  IFNet can successfully combat handheld motion errors. In contrast, although SRGAN \cite{srgan} and AMGAN \cite{amgan} effectively reduce the defocus effect, they fail to accurately restore the shapes of unseen objects.}
	\label{qual_letter}
\end{figure*}

\subsubsection{Simulated dataset with 300 real-life objects}
 Collecting SAR signals from a large number of real-life and diverse objects is challenging. Therefore, to evaluate the performance of IFNet on various complex objects, we simulate SAR signals using 3D real-life models in the ShapeNet \cite{shapenet} dataset, as shown in Fig. \ref{shapenet_dataset}. ShapeNet is a large-scale 3D model dataset consisting of 55 common object categories with about 51,300 unique 3D models. We choose 300 models covering knives, pistols, and rifles with different shapes and sizes. The simulated SAR signal of each model is distorted by 50 distinct handheld motion errors, resulting in a total of 15,000 images. To evaluate the ability of IFNet to generalize across unseen objects, we randomly select 10,500 images from 210 models for training and test the rest 4,500 images from 90 new models. 

 The SAR signal simulation has the following configurations: the signal frequency is 77 GHz; the dimension of the planar array is $256 \times 128$; the distance between adjacent virtual elements along the x-axis and y-axis is 1 mm and 1.9 mm, respectively; the distance between the target and the array plane is 0.3 m; the size of the imaging scene is about 260 mm$ \times$ 240 mm. Since our focus is to test the performance on a large number of real-life objects, we do not simulate complex radio propagation such as multipath and environmental reflections to achieve a more efficient implementation.

\subsection{Evaluation with Real Measurements}
To demonstrate the performance of IFNet on different handheld scanning patterns and object shapes, we present the results of IFNet and baselines when evaluating with real measurements. Note that in this section, we set both the number of stages and the number of Resblocks in each stage as 10, which is different from the settings in our previous conference paper (7 for the number of stages and 6 for the number of ResBlocks in each stage) \cite{ifnet}. This is because we find that a deeper network with more trainable parameters usually results in stronger learning capability and better performance.

\begin{figure*}[htbp]
	\centering
	\includegraphics[scale=0.6]{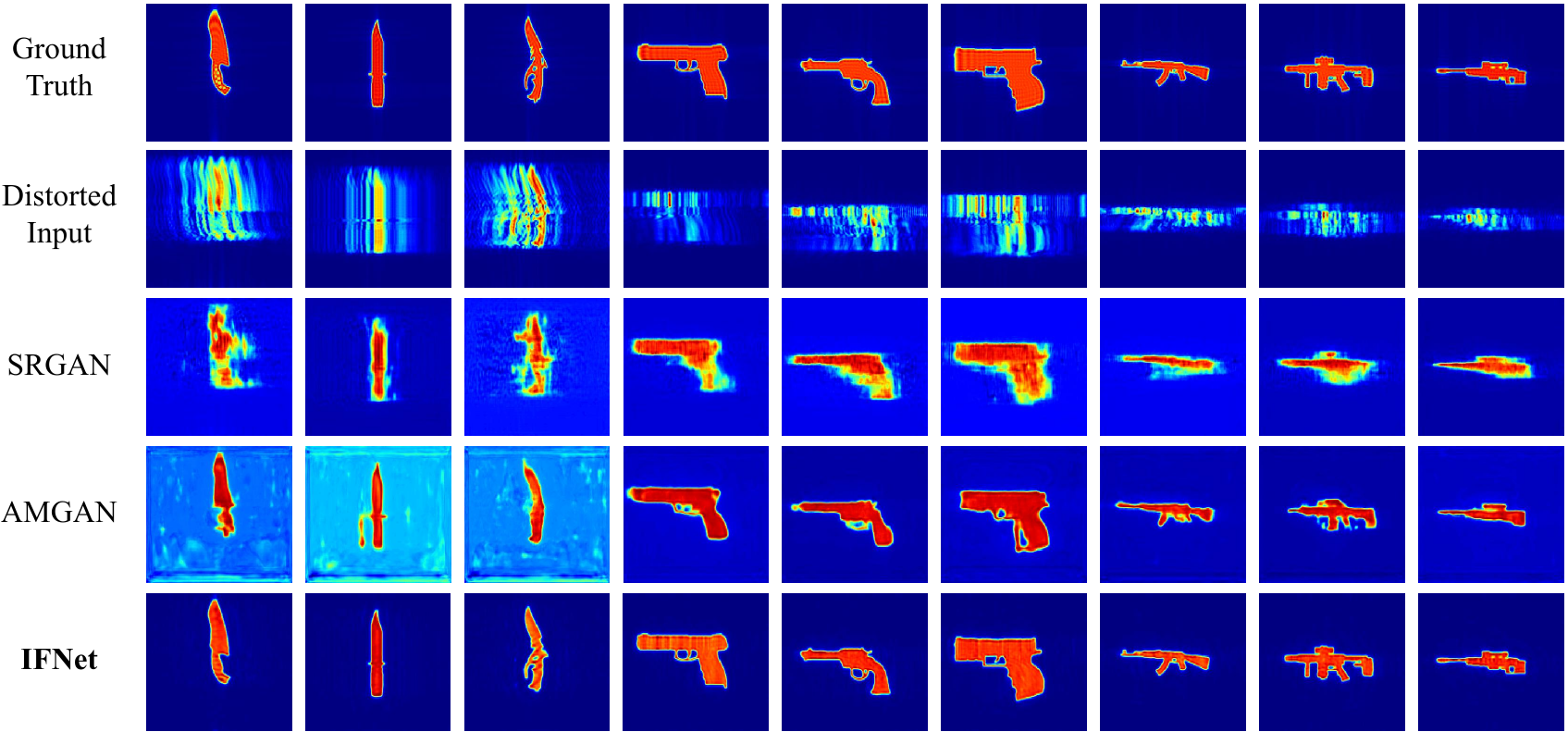}
	\caption{Qualitative comparison with various real-life objects from the simulated dataset. The testing objects are not included in the training set. IFNet performs best among all methods and reconstructs high-fidelity images.}
	\label{qual_shapenet}
\end{figure*}

\subsubsection{Qualitative comparison}
The qualitative comparison between IFNet and baselines with real measurements is presented in Fig. \ref{qual_letter}. The results demonstrate that the quality of images generated by IFNet surpasses that of cGAN-based approaches. 
Specifically, the results on IFNet show high similarity with the ground-truth images: the defocus effect is eliminated, and the shapes and details are well-reconstructed. In contrast, while SRGAN and AMGAN can mitigate the defocus effect to some extent, they struggle to generate accurate target shapes due to limited generalization. For instance, the letters H, W, and X output by AMGAN are more like the letters R, N, and Y, respectively. On the other hand, the letter O can be better restored because it is similar to the letter Q, which is in the training set. These results show that AMGAN can only cope with similar objects and fails to generalize to new shapes not seen during training. This is because these methods directly borrow from RGB image processing networks and ignore phase information. Conversely, the design of IFNet is based on an interpretable handheld imaging optimization model, which takes into account the characteristics of mmWave images and handheld phase errors. Moreover, the network input also includes phase information, which enhances its generalization capability for imaging different targets.

% Conversely, since IFNet is constructed based on a signal model that exploits the characteristics of handheld phase errors, it can better capture the correlations between handheld phase errors and imaging performance. 

\subsubsection{Quantitative comparison}
\begin{figure}[htbp]
	\centering
	\includegraphics[scale=0.45]{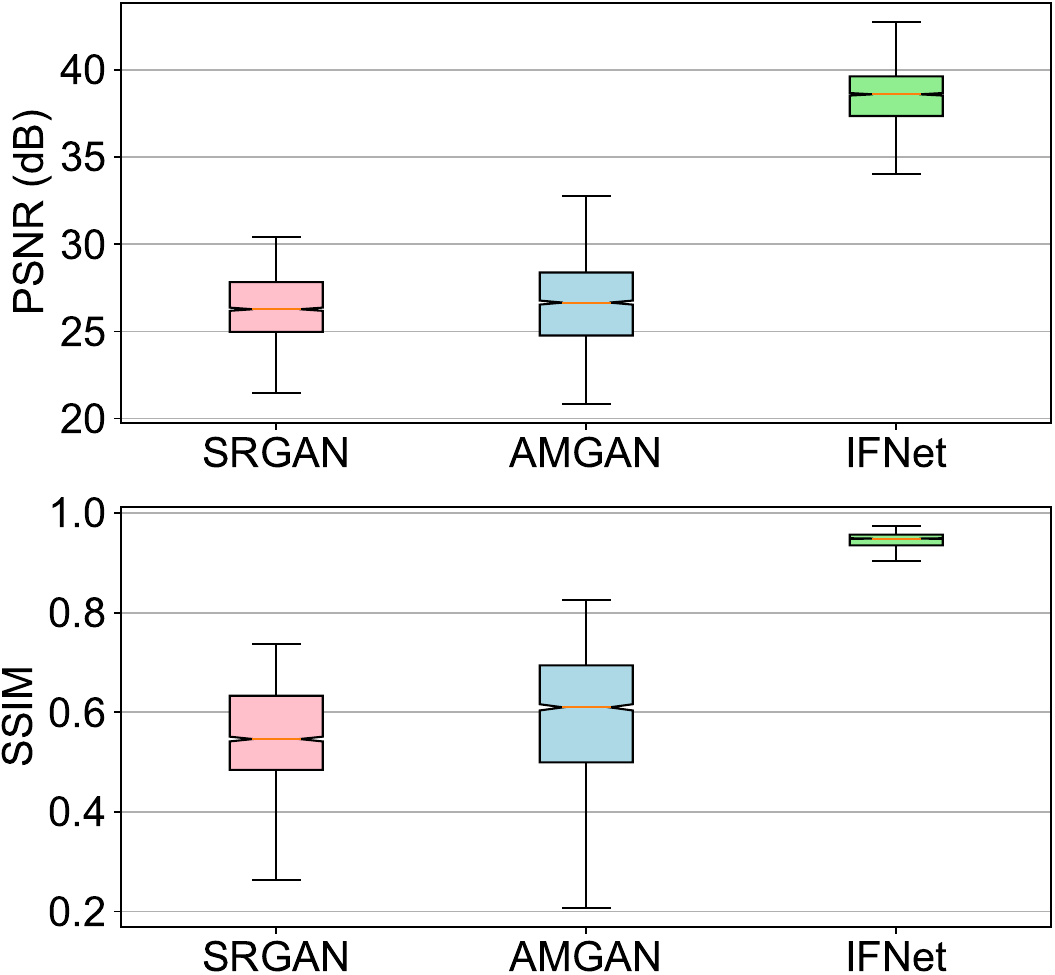}
	\caption{Qualitative comparison with 26 letters from real measurements.}
	\label{quant_letter}
\end{figure}

For a quantitative comparison between IFNet and baselines, we present the average PSNR and SSIM values of all testing images generated by IFNet, SRGAN, and AMGAN in Fig. \ref{quant_letter}. It is evident that IFNet significantly outperforms those cGAN-based methods in both metrics. Specifically, the average PSNR and SSIM values of images produced by IFNet are 38.28 dB and 0.94, respectively, while for SRGAN/AMGAN, the PSNR and SSIM values are 26.39 dB/26.12 dB and 0.57/0.53, respectively.  In other words, IFNet achieves at least 11.89 dB improvement in average PSNR and 64.91\% improvement in
average SSIM. This demonstrates that IFNet effectively mitigates the distortion caused by handheld scanning and successfully reconstructs SAR images with high visual fidelity. Again, this is because IFNet models the estimation and compensation of handheld phase errors, allowing it to better learn the distribution of phase errors and achieve improved imaging results on different objects. However, SRGAN and AMGAN only follow the approach of super-resolution in RGB images to tackle this problem, which prevents the network from learning the essential characteristics of phase errors. As a result, they struggle to generalize to different targets and reconstruct the accurate shape of objects.

% Table generated by Excel2LaTeX from sheet 'Sheet1'
\begin{table}[htbp]
  \centering
  \caption{Quantitative comparison with various real-life objects from the simulated dataset.}
\renewcommand\arraystretch{1.5}
\tabcolsep=0.5cm

    \begin{tabular}{lll}
    \hline
    Method & PSNR (dB)  & SSIM \\ \hline
    Input & 21.94 $\pm$ 2.70 & 0.82 $\pm$ 0.08 \\
    SRGAN \cite{srgan} & 26.39 $\pm$ 2.37 & 0.91 $\pm$ 0.03 \\
    AMGAN \cite{amgan} & 25.20 $\pm$ 3.03 & 0.88 $\pm$ 0.07 \\
    \textbf{IFNet} & \textbf{40.57} $\pm$ \textbf{3.67} & \textbf{0.98} $\pm$ \textbf{0.01} \\ \hline
    \end{tabular}%
    
  \label{quant_shapenet}%
\end{table}%

\subsection{Evaluation with Simulated Targets}
To investigate the performance of IFNet on various complex objects, we conduct experiments with simulated SAR signals from the ShapeNet dataset. While the simulated signals might be ideal due to the lack of multipath and reflections from the environment, they provide an efficient approach to measure IFNet's ability to recover a large number of real-life objects.  The
number of stages and the number of Resblocks in each stage
are set as 5 and 8, respectively. Note that the testing objects are all unseen targets. From the qualitative comparison illustrated in Fig. \ref{qual_shapenet}, we can see that IFNet excels in recovering fine-grained details of various complex objects from severely distorted handheld SAR images. The quantitative comparison shown in Table \ref{quant_shapenet} demonstrates that the PSNR and SSIM of images output by IFNet significantly surpasses that of SRGAN and AMGAN. These results indicate that IFNet can better capture the inherent characteristic of handheld phase errors rather than simply restoring the images.

\subsection{Ablation Study}
To further investigate the impact of different modules in IFNet on imaging performance, we conduct a series of ablation experiments to compare the imaging results with different design choices of IFNet. In these experiments, we use the dataset from real measurements for training and testing, and the training setting and strategy remain consistent with the previously mentioned details.

\subsubsection{Impact of the number of stages}
\begin{figure}[htbp]
	\centering
	\includegraphics[scale=0.42]{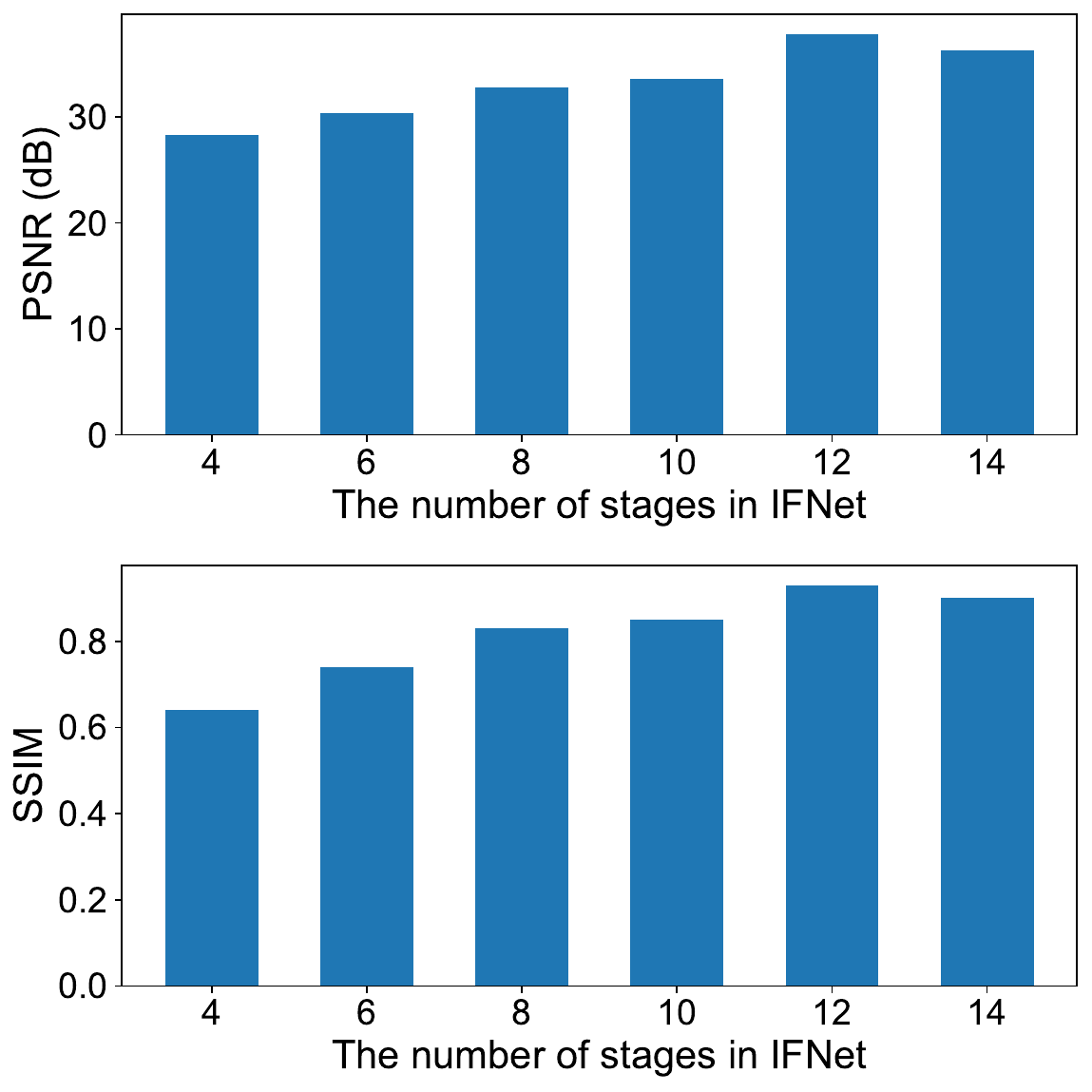}
	\caption{The PSNR and SSIM for a different number of stages in IFNet.}
	\label{stages}
\end{figure}

IFNet is composed of multiple imaging modules and focusing modules. Each combination of an imaging module and a focusing module is called a stage, corresponding to the update step of the iterative optimization algorithm. To explore the influence of different numbers of imaging and focusing modules on the final results, this study sets the number of ResBlocks in the focusing module to 6 and gradually increases the number of imaging and focusing stages from 4 to 14. The PSNR and SSIM on the testing images are shown in Fig. \ref{stages}. It can be observed that as the number of stages increases, both PSNR and SSIM show an upward trend. However, when the number of stages reaches a certain quantity, the network performance no longer improves further and may even slightly decline. This is because increasing the number of modules enhances the network's ability to learn the distribution of handheld phase errors as the network depth increases. However, when the network becomes too deep, issues such as overfitting are more likely to occur, leading to a decrease in generalization on the test set.

\subsubsection{Impact of the number of ResBlocks}
\begin{figure}[htbp]
	\centering
	\includegraphics[scale=0.42]{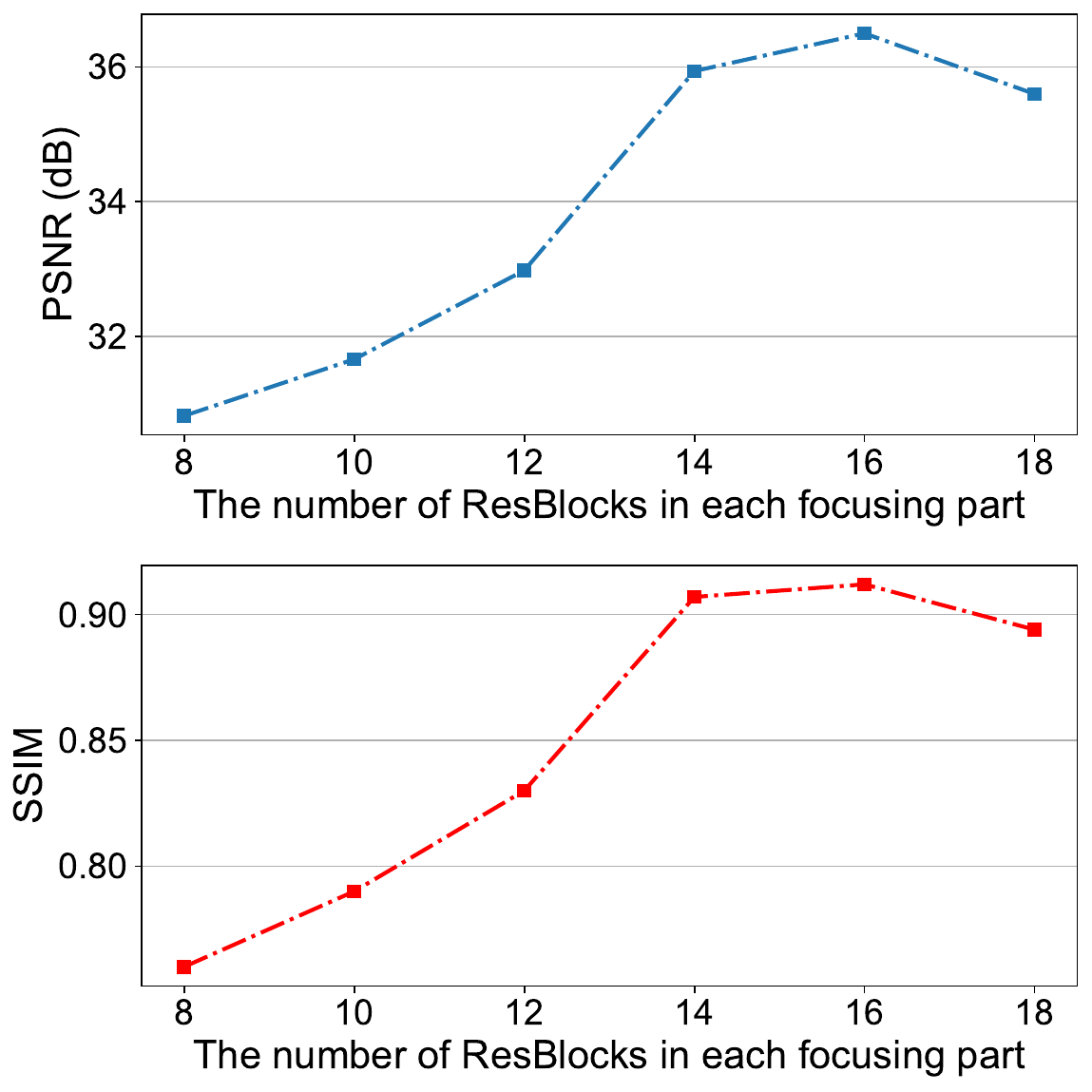}
	\caption{The PSNR and SSIM for a different number of ResBlocks in IFNet.}
	\label{resblock}
\end{figure}

Another approach to enhance the learning ability of IFNet is modifying the design of each module. Since the focusing module consists of multiple ResBlocks, we conduct a series of experiments to investigate the impact of different numbers of ResBlocks within the focusing module on handheld imaging results. Specifically, the number of imaging and focusing stages in IFNet is fixed at 5, while the number of ResBlocks within each focusing module gradually increases from 8 to 18. Fig. \ref{resblock} shows the variation of PSNR and SSIM for handheld imaging results with different numbers of ResBlocks. We can find that, similar to the previous findings, both PSNR and SSIM gradually increase as the number of ResBlocks increases. However, when the number of ResBlocks reaches a certain quantity, the imaging performance no longer improves and may even decline. Similarly, this is because deeper networks are better at learning the distribution characteristics of handheld phase errors. However, excessively deep networks can compromise generalization on the test set.

\section{conclusion}\label{conclusion}
In this paper, we introduced IFNet, a deep unfolding network that combines signal processing models and deep neural networks to address the challenges of handheld mmWave imaging. By integrating multiple priors and employing an iterative network structure, IFNet effectively compensated for phase errors and recovered high-fidelity images from severely distorted signals. Extensive experiments demonstrated the superior performance of IFNet compared to state-of-the-art methods both quantitatively and qualitatively. We believe that the proposed IFNet has the potential to inspire future works exploring innovative applications of learning-based methods in signal processing fields, thus breaking the performance bottlenecks and opening new avenues for practical applications.

\newpage
% References should be produced using the bibtex program from suitable
% BiBTeX files (here: strings, refs, manuals). The IEEEbib.bst bibliography
% style file from IEEE produces unsorted bibliography list.
% -------------------------------------------------------------------------

\bibliographystyle{IEEEbib}
\bibliography{ref}

\end{document}